\documentclass[runningheads]{llncs}

\usepackage{accvabbrv}

\usepackage{graphicx}
\usepackage{booktabs}

\usepackage[table]{xcolor}

\usepackage[accsupp]{axessibility}  

\usepackage{orcidlink}

\begin{document}

\title{ControlCol: Controllability in Automatic Speaker Video Colorization} 

\titlerunning{ControlCol}

\author{Rory Ward\inst{1,2}\orcidlink{0009-0003-7634-9946} \and
John G. Breslin\inst{1,2}\orcidlink{0000-0001-5790-050X} \and
Peter Corcoran\inst{1,2}\orcidlink{0000-0003-1670-4793}}

\authorrunning{R.Ward et al.}

\institute{SFI Centre for Research Training in Artificial Intelligence, Data Science Institute, University of Galway, University Road, H91 TK33, Ireland.  \and
School of Engineering, University of Galway, University Road, Galway, H91 TK33, Ireland}

\maketitle

\begin{abstract}
  Adding color to black-and-white speaker videos automatically is a highly desirable technique. It is an artistic process that requires interactivity with humans for the best results. Many existing automatic video colorization systems provide little opportunity for the user to guide the colorization process. In this work, we introduce a novel automatic speaker video colorization system which provides controllability to the user while also maintaining high colorization quality relative to state-of-the-art techniques. We name this system ControlCol. ControlCol performs 3.5\% better than the previous state-of-the-art DeOldify on the Grid and Lombard Grid datasets when PSNR, SSIM, FID and FVD are used as metrics. This result is also supported by our human evaluation, where in a head-to-head comparison, ControlCol is preferred 90\% of the time to DeOldify. Example videos can be seen in the supplementary material.
\end{abstract}

\section{Introduction}\label{sec:introduction}

Adding color to black-and-white images and videos is a slow and time-consuming process \cite{Pierre2021}. To achieve results that appear plausible to a human, attention must be paid to the minutest of details. Attempts have been made to automate this process, but human-level automatic colorization is still unattainable \cite{10.1007/978-3-031-20071-7_1,kim2022bigcolor,https://doi.org/10.48550/arxiv.2111.05826,liu2023video,https://doi.org/10.48550/arxiv.2203.17276,10.1007/978-3-031-25069-9_41}. For images, the colorization must only be consistent spatially in the image, whereas in videos, the colorization must also be stable throughout time. In addition to the temporal consistency requirement, there is also a need for controllability in automatic video colorization. This is because colorization is an inherently artistic operation, as multiple possible color interpretations exist for any given grayscale media. That said, some colorizations are deemed more appropriate than others. For instance, human faces are under particular scrutiny due to their relevance in conveying non-verbal information \cite{ERICKSON200352}. In recognition of this, we choose to focus on speaker video colorization.

The main contributions of this work are as follows:

\begin{itemize}
    \item We introduce a novel, controllable, temporally consistent automatic speaker video colorization system. Which we name ControlCol (Ours).
    \item The proposed approach achieves state-of-the-art results on the Grid \cite{cooke_martin} and Lombard Grid \cite{lombardGrid} datasets. It has an average of 7\% performance gain over the previous state-of-the-art DeOldify \cite{antic2019deoldify} on the Grid \cite{cooke_martin} dataset and 3.5\% on the combination of the Grid \cite{cooke_martin} and Lombard Grid \cite{lombardGrid} datasets. When PSNR \cite{5596999}, SSIM \cite{1284395}, FID \cite{heusel2018gans} and FVD \cite{unterthiner2019fvd} are used as the metrics. 
    \item ControlCol (Ours) is also preferred to the previous state-of-the-art DeOldify \cite{antic2019deoldify} 90\% of the time in our user survey. In the same study, we show that there is only a 1\% preference for the ground truth over the proposed approach on the Lombard Grid \cite{lombardGrid} dataset.
\end{itemize}

\section{Related Work}\label{sec:related_work}

In this work, we decompose the challenging task of controllable automatic speaker video colorization into two more manageable sub-tasks. The first sub-task implements interactive image colorization to allow for controllability. The second extends the automatic image colorization task to the automatic video colorization domain. Therefore, this related work section will examine the literature surrounding automatic and interactive colorization for images and videos.

\subsection{Automatic Image Colorization}

Automatic image colorization is a well-established field \cite{cao2023animediffusion,vitoria2020chromagan,liang2024control,kang2023ddcolor,10.1145/3550454.3555432}. Some approaches from this category of colorizers that we are going to compare against are DeOldify \cite{antic2019deoldify}, ``Colorization Transformer'' (ColTran) \cite{https://doi.org/10.48550/arxiv.2102.04432}, and ``Generative Color Prior'' (GCP) \cite{wu2022vivid}. DeOldify \cite{antic2019deoldify} is a self-attention generative adversarial network-based (SAGAN) colorizer \cite{zhang2019selfattention} trained with a two time-scale update rule \cite{heusel2018gans}. ColTran \cite{https://doi.org/10.48550/arxiv.2102.04432} is a transformer-based \cite{https://doi.org/10.48550/arxiv.1706.03762} colorizer. It colorizes in three steps. It first uses a conditional autoregressive transformer to produce a low-resolution coarse coloring of the grayscale image. It then upsamples the color information to a higher number of bits. It finally upsamples the spatial data to get the desired resolution. GCP \cite{wu2022vivid} is a GAN-based colorizer. It leverages generative color priors by using ``retrieved'' matched features via a GAN encoder.

\subsection{Interactive Image Colorization}


\subsubsection{Exemplar Guided Colorization}

A common way of conditioning the colorization process is to give the model context through the form of an exemplar image \cite{9156389,10.1145/3394171.3413594,he2018progressive}. The machine learning model can take influence from the exemplar image's constituent colors when colorizing the black-and-white image. A difficulty with this approach is that the quality of the exemplar provided dramatically impacts the quality of the final colorization \cite{6655920,10.1145/2669024.2669037}.

\subsubsection{Text Guided Colorization}

Another intuitive way humans interact with machine learning models is through text \cite{nokey,lai2023minidalle3}. This is also true in interactive image colorization where work is being done to condition the colorization process with text \cite{RN162,Weng_Wu_Chang_Tang_Li_Shi_2022,10204477}. One of the main challenges associated with this approach is bridging the difference in modalities of the structured text to the unstructured pixels \cite{chen2018languagebased,10.1145/3355089.3356561}. The state-of-the-art technique in this area is ``Language-based Colorization with Any-level Descriptions using Diffusion Priors'' (L-CAD) \cite{chang2023lcad}. It is a latent-diffusion-based \cite{nokey} automatic image colorization system which colorizes with natural language descriptions of any granularity. It provides instance-aware colorization across a range of domains.

\subsection{Automatic Video Colorization}

One of the simplest ways of performing automatic video colorization is to perform automatic image colorization on each video frame independently, but this results in poor temporal consistency and, therefore, poor quality overall. More sophisticated systems take temporal consistency into account, such as ``Video Colorization with Hybrid Generative Adversarial Network'' (VCGAN) \cite{Zhao_2023} and Ward and Breslin \cite{imvi2022}. VCGAN \cite{Zhao_2023} is a recurrent video colorization system with a generative adversarial network-based model \cite{goodfellow2014generative}. It has temporal consistency incorporated by design through feed-forward feature extractors and a temporal consistency loss. Ward and Breslin \cite{imvi2022} propose a hybrid deep-learning and exemplar-based approach. It ensures temporal consistency through an exemplar-based automatic video colorization system.

\subsection{Interactive Video Colorization}

Some interactive video colorization systems also exist but are generally less widespread because it is a more challenging process \cite{he2018deep,yang2022bistnetsemanticimageprior,9693178}. DeepRemaster \cite{IizukaSIGGRAPHASIA2019} is one of the most prominent methods using exemplar conditioning to control the system. Its underlying model is a temporal convolutional neural network with attention mechanisms. It is trained on a dataset of videos that have been artificially deteriorated, including those that have been desaturated. In addition to colorization, it also performs super-resolution, noise removal, and contrast enhancement.

\section{Methodology}\label{sec:methodology}

\begin{figure*}[h]
  \centering
  \includegraphics[height=0.82\linewidth]{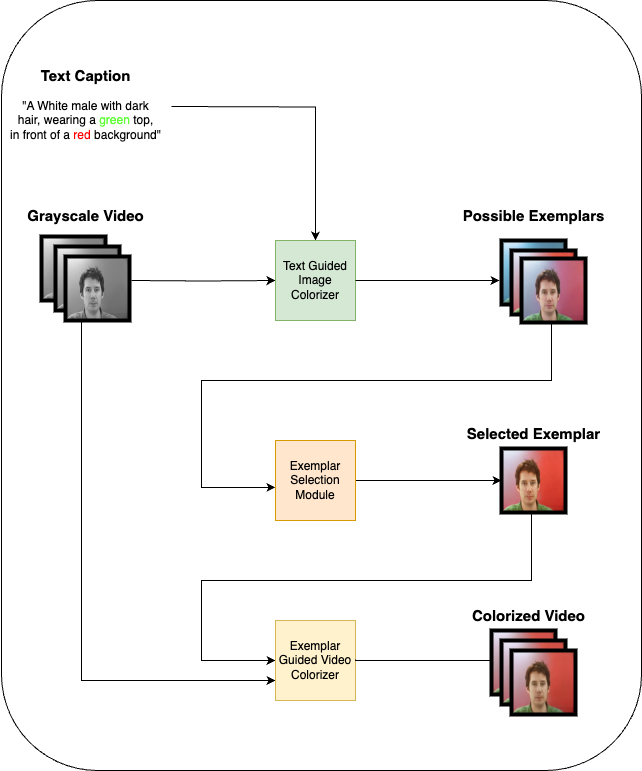}
  \caption{\textbf{System Architecture} of our proposed method. ControlCol (Ours) takes a grayscale video and a text caption as input. It produces a temporally consistent colorized video as output. A text-guided image colorizer, exemplar selection module and an exemplar-guided video colorizer are used in the system.}
  \label{fig:Architecture}
\end{figure*}

\subsection{Data Selection}

As we are primarily focused on speaker video colorization we chose to use the Grid \cite{cooke_martin} and Lombard Grid \cite{lombardGrid} datasets. The Grid Corpus \cite{cooke_martin} consists of high-quality audio and video (facial) recordings of 1000 sentences spoken by each of the 34 talkers (18 male, 16 female). The Lombard Grid \cite{lombardGrid} dataset consists of 54 talkers (30 male, 24 female), with 100 utterances per talker. 

\subsection{Data Pre-Processing}

The data consisted of color videos at 360×288 pixels without text captions. Therefore, the dataset needed to be desaturated to create pairs of color and grayscale images. The data was at an unorthodox resolution and needed to be rescaled to a more standard quality of 128x128 pixels. The dataset did not come with text captions by default, so these text captions were required to be generated. There were too many frames to do this manually, so the automatic image captioning capabilities of  GPT4 \cite{openai2024gpt4} were employed for this task.

\subsection{ControlCol System Architecture}

The system architecture can be seen in Fig.~\ref{fig:Architecture}. The system takes the grayscale video and a text caption to guide the colorization as input. Both are passed to the text-guided image colorization system, namely L-CAD \cite{chang2023lcad}. From the text-guided image colorization system, a series of possible exemplars are generated, each with varying levels of quality. These possible exemplars $p\in P$ are then ranked by the exemplar selection module, which will select the most suitable exemplar $E$. Within the exemplar selection module, the minimum $min()$ normalized ``Face Image Quality'' (FIQ) \cite{terhörst2020serfiqunsupervisedestimationface} $norm(FIQ(p))$  score is calculated, and the associated exemplar extracted, as shown in Eq.~\eqref{eqn:Exemplar_Selection}. This module is inspired by Ward and Breslin \cite{imvi2022}, but instead of the combination of the ``natural image quality evaluator'' (NIQE) \cite{6353522} and ``blind/referenceless image spatial quality evaluator'' (BRISQUE) \cite{6272356} score that they use, we use FIQ \cite{terhörst2020serfiqunsupervisedestimationface}. FIQ \cite{terhörst2020serfiqunsupervisedestimationface} is used as it is more specifically used for faces than BRISQUE \cite{6272356} and NIQE \cite{6353522}. This chosen exemplar guides the temporally consistent video colorization (DeepRemaster \cite{IizukaSIGGRAPHASIA2019}), taking the grayscale video as input. This results in the final output of the system, which is the temporally consistent text-guided colorized video. 

\begin{equation}
\label{eqn:Exemplar_Selection}
E = min(norm(FIQ(p))) \forall  p\in P
\end{equation}

\section{Evaluation}\label{sec:evaluation}

\subsection{Metrics}

Evaluating colorization performance is difficult as it is highly subjective \cite{https://doi.org/10.1002/col.22593}, but we also require an objective assessment of the relative approaches. Therefore, a selection of metrics has been employed. They are PSNR (Peak Signal to Noise Ratio) \cite{5596999}, SSIM (Structural Similarity Index) \cite{1284395}, FID (Fréchet inception distance) \cite{heusel2018gans}, and FVD (Fréchet video distance) \cite{unterthiner2019fvd}. PSNR \cite{5596999} is a per-pixel comparison between the target and reference images. It is a basic metric as it does not consider any semantic meaning in the images, simply their absolute differences. A higher PSNR \cite{5596999} indicates better quality. SSIM \cite{1284395} is a more sophisticated metric than PSNR \cite{5596999}. It calculates the differences in the structures of the objects found in the source and target images. A high SSIM \cite{1284395} value indicates better results. FID \cite{heusel2018gans} advances on SSIM \cite{1284395} and PSNR \cite{5596999} in that it takes account of all the images in a dataset and calculates the differences in the distribution as opposed to in the individual images. A lower FID \cite{heusel2018gans} represents better results. FVD \cite{unterthiner2019fvd} considers temporal consistency and differences in the distributions of the target and source images. A lower FVD \cite{unterthiner2019fvd} indicates more consistent colorizations. MOS (Mean Opinion Score) \cite{HUANG2022105006} measures the average perceived quality of the colorizations. We calculate it as the sum of the responses divided by the number of options. A higher MOS \cite{HUANG2022105006} indicates a better opinion of the colorizations.

\subsection{Quantitative}

\begin{table*}[t]
\centering
\begin{tabular}{cccccc}
\\ 
\hline
                    
Dataset & Method              & PSNR ↑ & SSIM ↑ & FID ↓ & FVD ↓ \\ \hline
Grid \cite{cooke_martin}& DeOldify \cite{antic2019deoldify}            & 28.07  & 0.79   & 52.67  & 520.75  \\
& DeepRemaster \cite{IizukaSIGGRAPHASIA2019}        & 27.70  & 0.77   & 108.68 & 927.91 \\
  & ColTran \cite{https://doi.org/10.48550/arxiv.2102.04432}            & 28.08  & 0.76    & 91.76 & 759.32 \\
& GCP \cite{wu2022vivid} & 27.74 & 0.75 & 109.75 & 1555.53\\
& VCGAN \cite{Zhao_2023} & 27.86 & 0.83 & 67.79 & 951.28\\ 
& L-CAD \cite{chang2023lcad} & \textbf{28.47} & \textbf{0.88} & 33.34 & 561.54\\
 & ControlCol (Ours) with BN & 27.91 & 0.87 & \textbf{26.23} & \textbf{399.62} \\ \hline
\cellcolor{yellow!25} & \cellcolor{yellow!25} ControlCol (Ours) & \cellcolor{yellow!25} 27.71 & \cellcolor{yellow!25} 0.85 & \cellcolor{yellow!25} 33.12 & \cellcolor{yellow!25} 467.52\\ \hline
Lombard Grid \cite{lombardGrid} & DeOldify \cite{antic2019deoldify}            &  30.30  & 0.93   & 28.27 & 520.94 \\
& DeepRemaster \cite{IizukaSIGGRAPHASIA2019}        & 30.09  & 0.95   & 32.9 & 1382.56 \\
& ColTran \cite{https://doi.org/10.48550/arxiv.2102.04432}            & 29.96  & 0.89    & 37.7 & 1583.94 \\
& GCP \cite{wu2022vivid} & 29.86 & 0.91 & 85.09 & \textbf{419.75} \\
& VCGAN \cite{Zhao_2023} & 30.20 & \textbf{0.96} & 72.17 & 2146.79 \\
& L-CAD \cite{chang2023lcad} & \textbf{30.68} & 0.95 & 31.4 & 640.85 \\
 & ControlCol (Ours) with BN & 30.55 & 0.94 & 40.08 & 1007.82\\ \hline	
\cellcolor{yellow!25} & \cellcolor{yellow!25} ControlCol (Ours) & \cellcolor{yellow!25} 30.64 & \cellcolor{yellow!25} 0.94 & \cellcolor{yellow!25} \textbf{26.26} & \cellcolor{yellow!25} 531.98 \\
 \hline  	
Overall & DeOldify \cite{antic2019deoldify}            & 29.19  & 0.86   & 40.47  & 520.85 \\
& DeepRemaster \cite{IizukaSIGGRAPHASIA2019}        & 28.90  &  0.86  & 81.08 & 1155.24 \\
& ColTran \cite{https://doi.org/10.48550/arxiv.2102.04432}            & 29.02  & 0.83 & 64.73 & 1171.63 \\
& GCP \cite{wu2022vivid} & 28.80 & 0.83 & 97.42 & 987.64 \\
& VCGAN \cite{Zhao_2023} & 29.03 & 0.90 & 64.98 & 1549.04 \\
& L-CAD \cite{chang2023lcad} & \textbf{29.58} & \textbf{0.92} & 32.37 & 601.20 \\
 & ControlCol (Ours) with BN & 29.23 & 0.91 & 33.16 & 703.72 \\ \hline	
\cellcolor{yellow!25} & \cellcolor{yellow!25} ControlCol (Ours) & \cellcolor{yellow!25} 29.18 & \cellcolor{yellow!25} 0.90 & \cellcolor{yellow!25} \textbf{29.69} & \cellcolor{yellow!25} \textbf{499.75} \\
 \hline  
\end{tabular}
\caption{\textbf{Quantitative analysis} of the colorization methods using the relevant metrics on the chosen datasets. ↑ and ↓ indicates the optimal score direction.}
\label{tab:Quantitative-Comparison}
\end{table*}

The results can be numerically compared in Tab.~\ref{tab:Quantitative-Comparison}. This table was constructed by colorizing a grayscale subset of the Gird \cite{cooke_martin} and Lombard Grid \cite{lombardGrid} datasets using each of the colorization methods. The combined scores from both datasets are also shown (Overall). These colorizations were then evaluated using the standard metrics of PSNR \cite{5596999}, SSIM \cite{1284395}, FID \cite{heusel2018gans}, and FVD \cite{unterthiner2019fvd}. The results of these tests were then tabulated. ControlCol (Ours) has an average of 7\% performance gain over the previous state-of-the-art DeOldify \cite{antic2019deoldify} on the Grid \cite{cooke_martin} dataset. On average of the Grid \cite{cooke_martin} and Lombard Grid \cite{lombardGrid} datasets, our proposed method performs 3.5\% better than DeOldify \cite{antic2019deoldify}.

\begin{figure*}[h]
  \centering
  \includegraphics[height=0.9\linewidth]{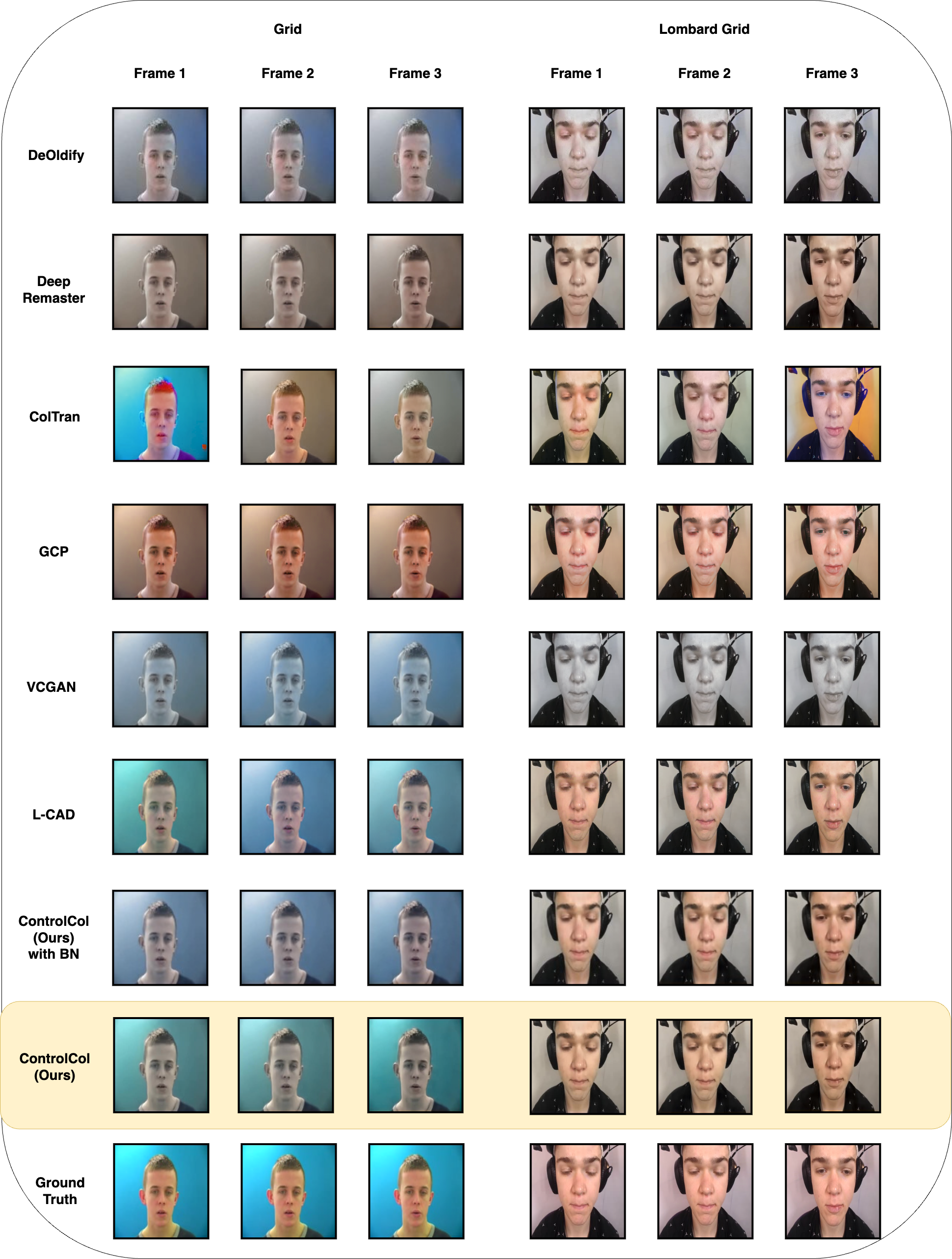}
  \caption{\textbf{Qualitiative analysis} of the colorization methods (DeOldify \cite{antic2019deoldify}, DeepRemaster \cite{IizukaSIGGRAPHASIA2019}, ColTran \cite{https://doi.org/10.48550/arxiv.2102.04432}, GCP \cite{wu2022vivid}, VCGAN \cite{Zhao_2023}, L-CAD \cite{chang2023lcad}, ControlCol (Ours) with BN (BRISQUE \cite{6272356} and NIQE \cite{6353522} exemplar selection) and ControlCol (Ours)) on the datasets (Grid \cite{cooke_martin}, and Lombard Grid \cite{lombardGrid}). The ground truth sequences are also provided for reference.}
  \label{fig:Video_Comparison}
\end{figure*}

\subsection{Qualitative}

The results can be visually compared in Fig.~\ref{fig:Video_Comparison}. This diagram was constructed by taking three consecutive frames from the datasets and colorizing them using each state-of-the-art colorizers chosen for comparison. The ground truth is also provided to give a reference point for what the output colorizations should look like. DeOldify \cite{antic2019deoldify} has produced less vibrant colorizations. The colorizations seem to have a halo around the subject and little contrast between the foreground and background. ColTran \cite{https://doi.org/10.48550/arxiv.2102.04432} has produced colorful outputs that are inconsistent throughout time. GCP \cite{wu2022vivid} has produced colorful outputs, which are also consistent over time. They are not accurate in terms of the ground truth colors. VCGAN \cite{Zhao_2023} has produced a poor colorization with a blue hue over the entire sequence. They are not accurate in terms of the ground truth colors. There is very little distinction between the subject and the background. ControlCol (Ours) has produced better colorizations than either L-CAD \cite{chang2023lcad} or DeepRemaster \cite{IizukaSIGGRAPHASIA2019} separately. The outputs are less drab than in DeepRemaster \cite{IizukaSIGGRAPHASIA2019} and are more consistent than in L-CAD \cite{chang2023lcad}. There is somewhat of a distinction between the foreground and the background, but there are still possible improvements.

\subsection{Controllablity}

\begin{figure*}[h]
  \centering
  \includegraphics[height=0.8\linewidth]{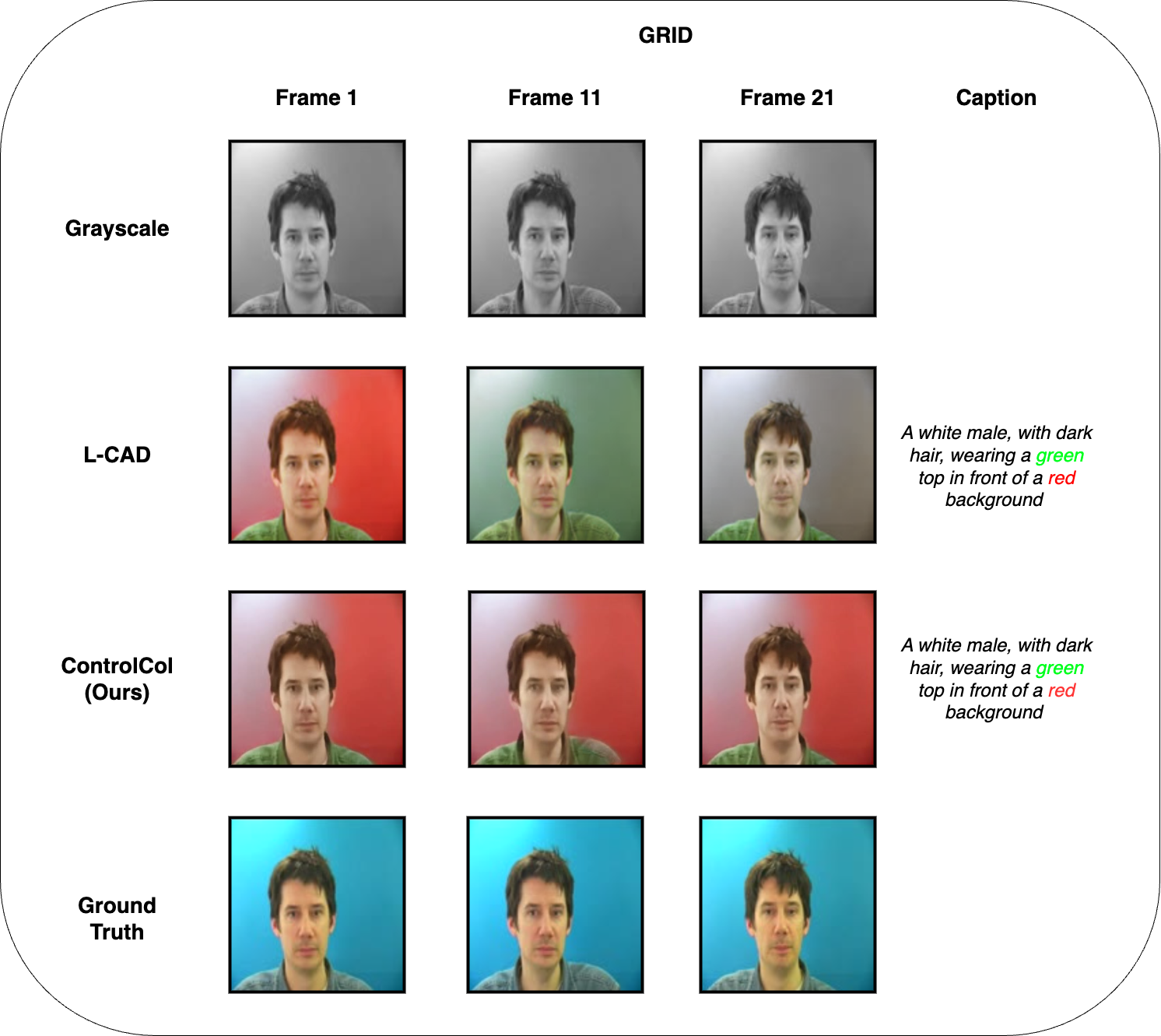}
  \caption{\textbf{Controllability analysis} of L-CAD \cite{chang2023lcad} and ConrolCol (Ours) on three frames taken from the Grid \cite{cooke_martin} dataset at ten frame intervals starting from the first frame. The video's grayscale and ground truth versions are also provided for reference. The caption is held constant for a fair comparison, ``A white male, with dark hair, wearing a green top in front of a red background''.}
  \label{fig:Controllability}
\end{figure*}

As controllability in automatic speaker video colorization is the focus of this paper, it must be evaluated as rigorously as the accuracy to the ground truth. This evaluation will be perfromed visually; see Fig.~\ref{fig:Controllability}. This figure was created by colorizing a subset of the dataset using two different text-based colorization methods. The grayscale and ground truth versions are also provided in the first and last row, respectively, for reference. The caption used to condition the colorization system is also provided in each case. The caption was held constant for each colorization to allow for a fair test. This caption is ``A white male, with dark hair, wearing a green top on front of a red background.'' This caption was chosen because it tested the model's ability to generalize to unseen requirements: a green top and a red background. The color of the hair was not changed as this could potentially conflict with the luminance values present in the grayscale frame. Samples were taken from the resultant colorizations at ten-frame intervals to determine the systems' abilities to generalize to longer-duration videos. The output of L-CAD \cite{chang2023lcad} (second row) is colorful but inconsistent throughout time. ControlCol (Ours) (third row) outputs colorful, consistent colorizations faithful to the text caption. It has managed to colorize the top green and achieve high contrast between the subject and the background.

\subsection{Survey}

\begin{figure*}[h]
  \centering
  \includegraphics[height=0.25\linewidth]{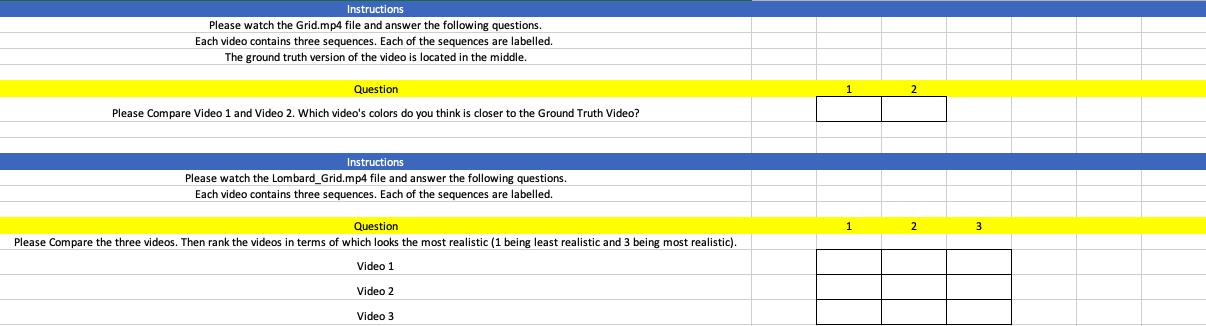}
  \caption{\textbf{The questions asked in the survey}. Before answering the questionnaire, the participants were shown two sets of three videos. They were then asked for their opinion on both sets of videos. This was then recorded on the questionnaire before being tallied and analysed.}
  \label{fig:Survey_Questions}
\end{figure*}

As colorization is such a subjective subject, the gold standard in evaluating automatic speaker video colorization systems is human evaluation \cite{HUANG2022105006}. Therefore, a survey was conducted to measure the relative performance between the colorizers. The survey has two questions, each relevant to a distinct dataset (Question 1 and Question 2). The two datasets are the Grid \cite{cooke_martin} and Lombard Grid \cite{lombardGrid} datasets. For each question, twenty-three 12-18 year old participants were questioned. None were color blind or had any visual impairment. See Fig.~\ref{fig:Survey_Questions} for the complete questionnaire. 

For Question 1, three versions of a clip selected from a subset of the Grid \cite{cooke_martin} dataset were shown simultaneously. One version of the clip was the ground truth video; another was the output of DeOldify \cite{antic2019deoldify}, and the third was the output of ControlCol (Ours). The ground truth video was labelled, but the other two were anonymized to reduce potential bias. DeOldify \cite{antic2019deoldify} and our proposed approach were labelled one and two to differentiate between them on the questionnaire. The participants were then asked to indicate which colorization they thought was closer to ground truth. This was a head-to-head comparison between our proposed method and the approach with the next best quantitative performance on the overall dataset using the FVD \cite{unterthiner2019fvd} metric. Namely DeOldify \cite{antic2019deoldify}. 

\begin{figure*}[h]
  \centering
  \includegraphics[height=0.7\linewidth]{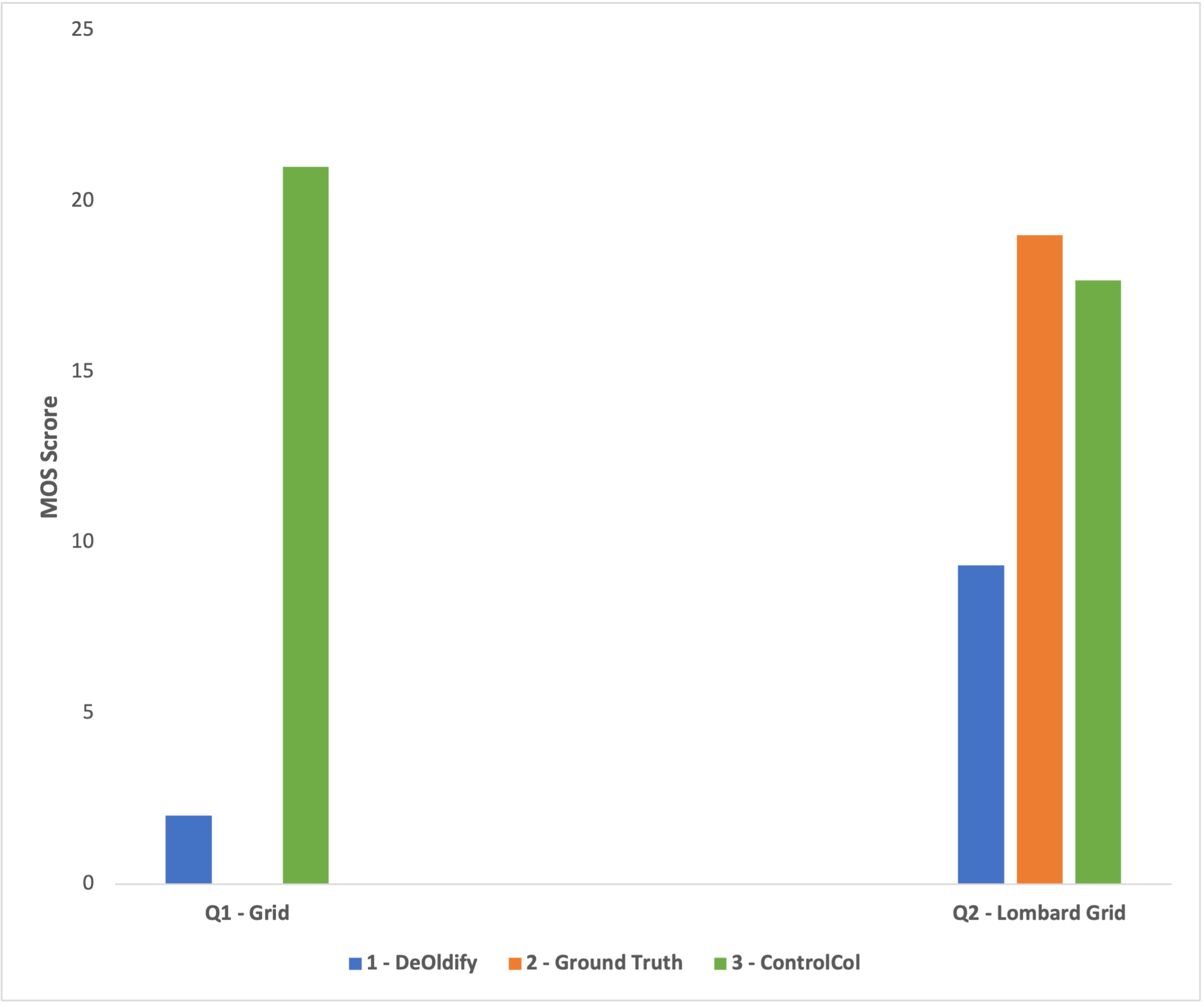}
  \caption{\textbf{Results of the survey.} The Y axis represents the MOS \cite{HUANG2022105006} score for each approach. The left cluster of bars refers to Question 1, which regards the Grid \cite{cooke_martin} dataset. The right cluster of bars refers to Question 2, which regards the Lombard Grid \cite{lombardGrid} dataset. Blue bars correspond to votes for DeOldify \cite{antic2019deoldify}, orange bars correspond to the ground truth videos, and green bars correspond to ControlCol (Ours).}
  \label{fig:Survey_Results}
\end{figure*}

Question 2 followed a similar format to Question 1. This time, the video was taken from a subset of the Lombard Grid \cite{lombardGrid} dataset. As well as the output of DeOldify \cite{antic2019deoldify} and our approach being anonymized, so was the ground truth, with the labelling going from 1 to 3 to account for the extra unlabelled video. The participants were then asked to rank their video preferences with particular attention to consistency and realism. This question was designed to act as a visual Turing test \cite{Turing1950-TURCMA}. The better the colorizations, the more challenging it should be for the participants to choose the ground truth video. The survey results can be seen in Fig.~\ref{fig:Survey_Results}. On Question 1, DeOldify \cite{antic2019deoldify} got a MOS \cite{HUANG2022105006} score of 2, whereas our proposed approach got a MOS \cite{HUANG2022105006} score of 21. On Question 2, DeOldify \cite{antic2019deoldify} got a MOS \cite{HUANG2022105006} score of 9, the ground truth video got a MOS \cite{HUANG2022105006} score of 19, and our proposed method got a MOS \cite{HUANG2022105006} score of 18. 

\begin{figure*}[h]
  \centering
  \includegraphics[height=0.8\linewidth]{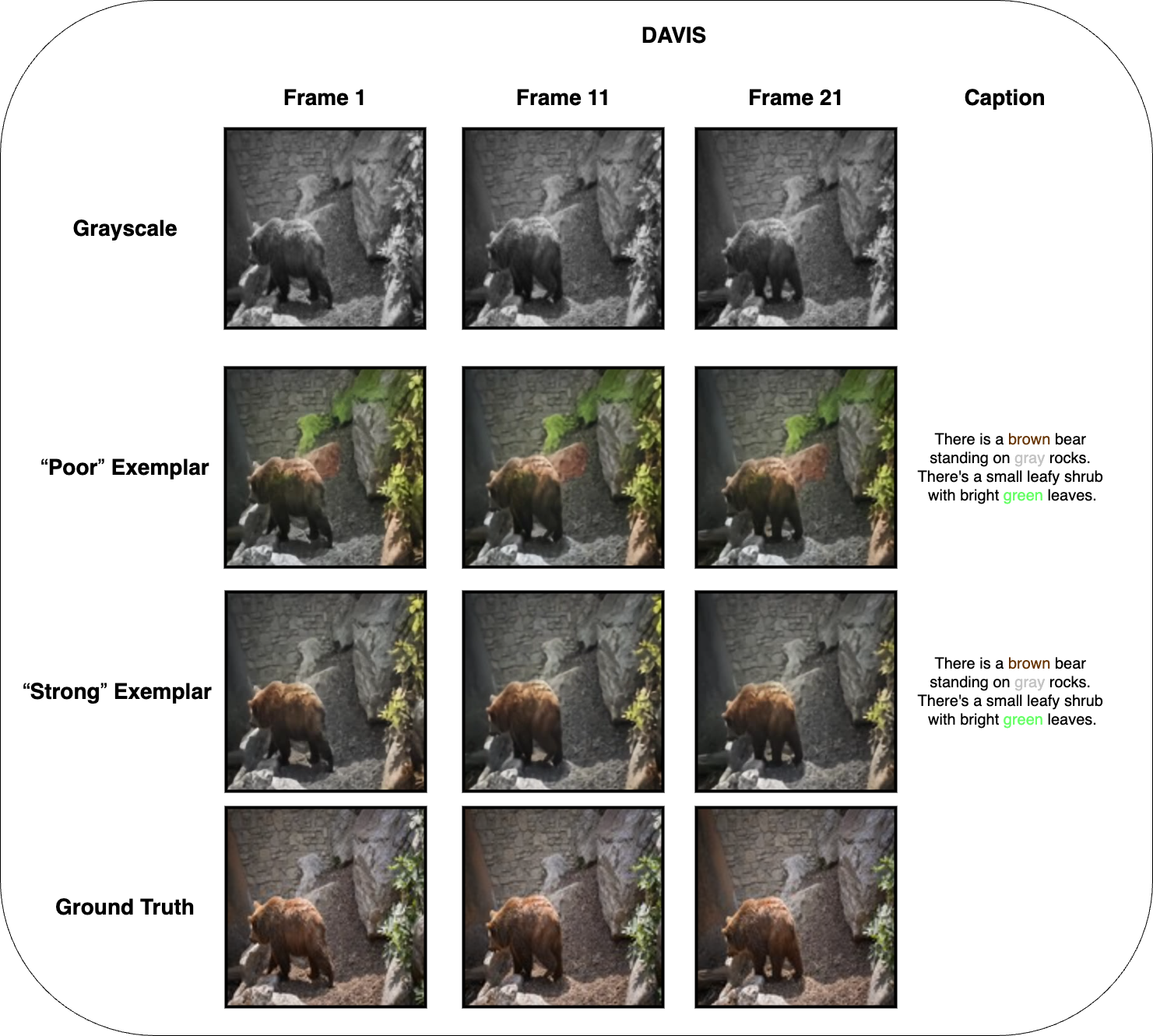}
  \caption{\textbf{Failure case} on a complex scene where SAM \cite{kirillov2023segment} struggled to segment the image fully, resulting in color bleeding. The exemplar selection module could not discern the ``poor'' exemplar from the ``strong'' exemplar. Three video sequence frames taken from the DAVIS \cite{Caelles_arXiv_2019} dataset are shown. The middle two rows have been colorized by ControlCol (Ours). The difference is that the sequence on the second row had a ``poor'' exemplar, and the sequence on the third row had a ``strong'' exemplar. The grayscale and ground truth are also given for reference. The caption is held constant for a fair test.}
  \label{fig:Failure_Cases}
\end{figure*}

\subsection{Failure Cases}

ControlCol (Ours) shows strong performance across various evaluations but struggles with certain aspects of automatic video colorization. One of the main challenges is difficulty adapting to out-of-domain test data. As our proposed approach leverages L-CAD \cite{chang2023lcad}, which uses SAM \cite{kirillov2023segment} for segmentation, it inherits the challenges associated with segmenting complex scenes; see Fig.~\ref{fig:Failure_Cases}. Three frames were taken from the DAVIS dataset \cite{Caelles_arXiv_2019} to provide our complex scene: Frame 1, Frame 11 and Frame 21. For reference, the frames' grayscale and ground truth versions are provided in the figure's top and bottom rows. The middle rows show the output of our system; the difference between the two is that the sequence in the second row from the top has been colorized with a ``poor'' exemplar, and the sequence on the third row from the top has been colorized with a ``strong'' exemplar, where ``strong'' and ``poor'' refer to how well the system was able to segment the image. As our method is particularly designed for speaker colorization it is unable to discern good exemplars from bad on out-of-domain data. Therefore, it chose the ``poor'' exemplar. To show what the system could achieve given the ``strong'' exemplar, it was provided with a ``strong'' exemplar by human selection. The caption was kept constant to allow for a fair comparison. The ``strong'' exemplar-guided colorization has produced an output consistent with the ground truth. The ``poor'' exemplar-guided colorization has produced an unrealistic colorization, propagating through the ``poor'' segmentation through the sequence, resulting in color bleeding between objects in the scene. Avoiding this situation relies on the strength of the exemplar selection module to discern the best exemplar.

\subsection{Ablation Study}

An ablation study was carried out to investigate the overall performance of each of the parts of the system, as shown in Fig.~\ref{fig:Video_Comparison} and Tab.~\ref{tab:Quantitative-Comparison}. The three aspects of the system being ablated are the controllability provided by L-CAD \cite{chang2023lcad}, the temporal consistency provided by DeepRemaster \cite{IizukaSIGGRAPHASIA2019}, and the interface between the two using the exemplar selection module. If the controllability provided by L-CAD \cite{chang2023lcad} is removed from the system, DeepRemaster \cite{IizukaSIGGRAPHASIA2019} without exemplar conditioning remains. The output of DeepRemaster \cite{IizukaSIGGRAPHASIA2019} without exemplar conditioning produces drab colorizations with muted tones and little contrast between the foreground and background. If the temporal consistency provided by DeepRemaster \cite{IizukaSIGGRAPHASIA2019} is removed from the system, L-CAD \cite{chang2023lcad} remains. The resultant colorizations are reasonable, but they are not consistent throughout time. An alternative exemplar selection module similar to Ward and Breslin \cite{imvi2022} was implemented to ablate the exemplar selection module. It uses the minimum combined BRISQUE \cite{6272356} and NIQE \cite{6353522} scores to select the exemplar. This is called ControlCol (Ours) with BN. 
When using the described metrics, ControlCol (Ours) with BN performs 3\% worse than ControlCol (Ours) on the overall dataset.

\section{Discussion}\label{sec:discussion}

\subsection{Quantitative}

Analyzing the overall results by metric, see Tab.~\ref{tab:Quantitative-Comparison}, ControlCol (Ours) performs equally to DeOldify \cite{antic2019deoldify} on PSNR \cite{5596999} and performs worse than L-CAD \cite{chang2023lcad} by 0.4\%. There is very little performance difference between the methods on this metric. PSNR \cite{5596999} measures the per-pixel difference in pairs of images and, therefore, is not a very robust colorization metric as human perception is more generally done at a higher level than pixel values. SSIM \cite{1284395} considers more object-level components and is, therefore, a better metric. On SSIM \cite{1284395}, our proposed method outperforms DeOldify \cite{antic2019deoldify} by 4\%, but is outperformed by L-CAD \cite{chang2023lcad} by 2\%. The issue with SSIM \cite{1284395} is that it compares images pair-wise and does not consider the overall distribution of the data. FID \cite{heusel2018gans} is a more useful metric than SSIM \cite{1284395} and PSNR \cite{5596999} as it considers the distribution of the data and not just pair-wise comparisons. On FID \cite{heusel2018gans}, ControlCol (Ours) outperforms DeOldify \cite{antic2019deoldify} by 11\% and L-CAD \cite{chang2023lcad} by 3\%. FID \cite{heusel2018gans} is a useful metric, but as it was originally designed for comparing images, it does not consider temporal consistency, unlike FVD \cite{unterthiner2019fvd}. FVD \cite{unterthiner2019fvd} is arguably the most relevant metric as it compares distributions and considers temporal consistency. On this metric, we achieve a 2\% increase in performance over DeOldify \cite{antic2019deoldify} and a 10\% increase over L-CAD \cite{chang2023lcad}. This large increase in performance over L-CAD \cite{chang2023lcad} and DeOldify \cite{antic2019deoldify} on this metric implies that ControlCol (Ours) can better ensure temporal consistency than the previous state-of-the-art techniques.

\subsection{Qualitative}

DeOldify's \cite{antic2019deoldify} less vibrant colorizations can be attributed to its GAN architecture, which is susceptible to mode collapse. This mode collapse in colorization results in muted tones. As ColTran \cite{https://doi.org/10.48550/arxiv.2102.04432} was not designed for automatic video colorization it is unable to keep its colorizations consistent through time. As GCP \cite{wu2022vivid} does not take additional conditioning, it cannot colorize with strong fidelity to the ground truth. VCGAN \cite{Zhao_2023} has performed poorly; it has a slight separation between the background and the foreground, indicating a poor semantic understanding of what is happening in the video sequence. ControlCol (Ours) has produced reasonable results relative to the prompt and the ground truth. It seems to have taken the positive aspects from each of the systems. It has taken the ability to be guided towards the correct colors through the use of L-CAD's controllability \cite{chang2023lcad} and combined that with DeepRemaster's ability to provide temporal consistency \cite{IizukaSIGGRAPHASIA2019}. There is still an over-reliance on blue, which has followed through from L-CAD \cite{chang2023lcad}, but again, this could be attributed to the quality of the text prompt and the number of segmentation masks per frame.

\subsection{Controllability}

Commenting on the results presented in Fig.~\ref{fig:Controllability}, we can say that the text conditioning has influenced each of the methods presented. L-CAD \cite{chang2023lcad} has initially been able to colorize the frames accurately, but it has not kept that consistent throughout the video sequence. This is to be expected as it does not have temporal consistency developed. An early hypothesis in this work was that by keeping the text caption consistent and understanding that there is a slight variance in the grayscale frames, temporal consistency should be enabled by default. However, this result proves this is not the case, and temporal consistency must be accounted for explicitly. ControlCol (Ours) has produced realistic colorizations consistent with the text prompt. There is a separation between the subject and the background, and the overall appearance is reasonable and colorful.

\subsection{Survey}

Analysing the results of the survey shown in Fig.~\ref{fig:Survey_Results}, ControlCol (Ours) receives greater than 90\% of the vote in the head-to-head comparison with DeOldify \cite{antic2019deoldify} on the Grid \cite{cooke_martin} dataset. This indicates a strong preference for our proposed approach on this dataset. On the Lombard Grid \cite{lombardGrid} dataset, DeOldify \cite{antic2019deoldify} receives 20\% of the vote, the ground truth receives 41\%, and our proposed approach receives 39\% of the vote.  This shows a strong preference for our approach relative to  DeOldify \cite{antic2019deoldify} (19\%). There is little difference between the percentage of the vote won by the ground truth and our proposed approach (2\%). This shows that on the Lombard Grid \cite{lombardGrid} dataset, our proposed approach is difficult to distinguish from the ground truth and, therefore, performant. 

\subsection{Failure Cases}

The selected exemplar's quality greatly impacts the overall colorization performance, as observed in Fig.~\ref{fig:Failure_Cases}. The exemplar selection module is specifically designed for speaker-centric videos; anything other than this is out of this project's scope. On out-of-domain data it cannot differentiate between a ``strong'' and ``poor'' exemplar. A ``poor'' exemplar with errors such as erroneous colors or color bleeding will propagate these errors throughout the entire video sequence.

\subsection{Ablation Study}

As DeepRemaster \cite{IizukaSIGGRAPHASIA2019} is primarily designed as an exemplar-based colorization technique, it cannot produce colorizations close to the ground truth without a suitable exemplar. L-CAD \cite{chang2023lcad} is also unable to keep its colorizations consistent throughout time. Choosing the correct exemplar greatly impacts the quality of the colorizations. Our exemplar selection module provides a robust way of doing this for faces, providing better results than using less specified metrics such as BRISQUE \cite{6272356} and NIQE \cite{6353522}. 

\section{Conclusion}\label{sec:conclusion}

In conclusion, we have introduced a novel controllable, temporally consistent automatic speaker video colorization system dubbed ControlCol (Ours). It achieves state-of-the-art results on the Grid \cite{cooke_martin} and Lombard Grid \cite{lombardGrid} datasets. The proposed method has an average of 7\% performance gain over the previous state-of-the-art DeOldify \cite{antic2019deoldify} on the Grid \cite{cooke_martin} dataset and an average of 3.5\% on the combination of the Grid \cite{cooke_martin} and Lombard Grid \cite{lombardGrid} datasets. This was measured using PSNR \cite{5596999}, SSIM \cite{1284395}, FID \cite{heusel2018gans} and FVD \cite{unterthiner2019fvd} as the metrics. It is also preferred to DeOldify \cite{antic2019deoldify} 90\% of the time in our user survey. In the same study, we show that there is only a 2\% preference for the ground truth over our proposed approach on the Lombard Grid \cite{lombardGrid} dataset. This work has confirmed that using text to control automatic video colorization is an intuitive mechanism for control and accuracy.

\section{Future Work}\label{sec:future_work}

To further develop this work, the number of masks per frame could be increased to investigate the effect of improved segmentation on colorization. It would also be interesting to investigate best practices for providing the most effective text prompt to this system. Adapting the system to work on out-of-domain data would also be useful for further developing this technology.

\bibliographystyle{splncs04}
\bibliography{main}
\end{document}